\def\year{2019}\relax
\documentclass[letterpaper]{article} 
\usepackage{aaai19}  
\usepackage{times}  
\usepackage{graphicx}
\usepackage{helvet}  
\usepackage{courier}  
\usepackage{url}  
\usepackage{graphicx}  
\usepackage{stfloats}
\usepackage{amsmath}
\usepackage{multirow}
\usepackage{booktabs}
\usepackage{diagbox}
\usepackage{makecell}
\usepackage{subfigure}
\begin{document}
%
\title{Cubic LSTMs for Video Prediction}
\author{Hehe Fan, Linchao Zhu, Yi Yang\\
Centre for Artificial Intelligence\\
University of Technology Sydney, Australia\\
\{hehe.fan,linchao.zhu\}@student.uts.edu.au, yi.yang@uts.edu.au\\
}
\maketitle
\begin{abstract}
Predicting future frames in videos has become a promising direction of research for both computer vision and robot learning communities. 
The core of this problem involves moving object capture and future motion prediction.
While object capture specifies which objects are moving in videos, motion prediction describes their future dynamics.
Motivated by this analysis, we propose a Cubic Long Short-Term Memory  (CubicLSTM) unit for video prediction. 
CubicLSTM consists of three branches, i.e., a spatial branch for capturing moving objects, a temporal branch for processing motions, and an output branch for combining the first two branches to generate predicted frames.
Stacking multiple CubicLSTM units along the spatial branch and output branch, and then evolving along the temporal branch can form a cubic recurrent neural network (CubicRNN). 
Experiment shows that CubicRNN produces more accurate video predictions than prior methods on both synthetic and real-world datasets.
\end{abstract}

\section{Introduction}
Videos contain a large amount of visual information in scenes as well as profound dynamic changes in motions.
Learning video representations, imagining motions and understanding scenes are fundamental missions in computer vision.
It is indisputable that a model, which is able to predict future frames after watching several context frames, has the ability to achieve these missions~\cite{DBLP:conf/icml/SrivastavaMS15}.
Compared with video recognition~\cite{DBLP:conf/cvpr/ZhuXY17,fan17complexeventdetection,fan18efficientvideoclassification}, video prediction has an innate advantage that it usually does not require external supervised information. 
Videos also provide a window for robots to understand the physical world.
Predicting what will happen in future can help robots to plan their actions and make decisions.
For example, action-conditional video prediction ~\cite{DBLP:conf/nips/OhGLLS15,DBLP:conf/nips/FinnGL16,DBLP:conf/corl/EbertFLL17,DBLP:conf/iclr/Babaeizadeh17} provides a physical understanding of the object in terms of the factors (e.g., forces) acting upon it and the long term effect (e.g., motions) of those factors.

As video is a kind of spatio-temporal sequences, recurrent neural networks (RNNs), especially Long Short-Term Memory (LSTM)~\cite{DBLP:journals/neco/HochreiterS97} and Gated Recurrent Unit (GRU)~\cite{DBLP:conf/emnlp/ChoMGBBSB14}, have been widely applied to video prediction.
One of the earliest RNN models for video prediction~\cite{DBLP:journals/corr/RanzatoSBMCC14} directly borrows a structure from the language modeling literature~\cite{DBLP:journals/jmlr/BengioDVJ03}.
It quantizes the space of frame patches as visual words and therefore can be seen as patch-level prediction. 
Later work~\cite{DBLP:conf/icml/SrivastavaMS15} builds encoder-decoder predictors using the fully connected LSTM (FC-LSTM). 
In addition to patch-level prediction, it also learns to predict future frames at the feature level. 
Since the traditional LSTM (or GRU) units learn from one-dimensional vectors where the feature representation is highly compact and the spatial information is lost, both of these methods attempt to avoid directly predicting future video frames at the image level. 
To predict spatio-temporal sequences, convolutional LSTM (ConvLSTM) \cite{DBLP:conf/nips/ShiCWYWW15} modifies FC-LSTM by taking three-dimensional tensors as the input and replacing fully connected layers by convolutional operations.
ConvLSTM has become an important component in several video prediction works~ \cite{DBLP:conf/nips/FinnGL16,DBLP:conf/nips/WangLWGY17,DBLP:conf/iclr/Babaeizadeh17,DBLP:conf/iclr/LotterKC17}.

Like most traditional LSTM units, FC-LSTM is designed to learn only one type of information, i.e., the dependency of sequences.
Directly adapting FC-LSTM makes it difficult for ConvLSTM to simultaneously exploit the temporal information and the spatial information in videos.
The convolutional operation in ConvLSTM has to process motions on one hand and capture moving objects on the other hand. 
Similarly, the state of ConvLSTM must be able to carry the motion information and object information at the same time.
It could be insufficient to use only one convolution and one state for spatio-temporal prediction.

In this paper, we propose a new unit for video prediction, i.e., the Cubic Long Short-Term Memory (CubicLSTM) unit.
This unit is equipped with two states, a temporal state and a spatial state, which are respectively generated by two independent convolutions.
The motivation is that different kinds of information should be processed and carried by different operations and states.
CubicLSTM consists of three branches, which are built along the three axes in the Cartesian coordinate system.
\begin{itemize}
    \item The temporal branch flows along the $x$-axis (temporal axis), on which the convolution aims to obtain and process motions. The temporal state is generated by this branch, which contains the motion information. 
    \item The spatial branch flows along the $z$-axis (spatial axis), on which the convolution is responsible for capturing and analyzing moving objects. The spatial state is generated by this branch, which carries the spatial layout information about moving objects.
    
    \item The output branch generates intermediate or final prediction frames along the $y$-axis (output axis), according to the predicted motions provided by the temporal branch and the moving object information provided by the spatial branch.
\end{itemize}
Stacking multiple CubicLSTM units along the spatial branch and output branch can form a two-dimensional network.
This two-dimensional network can further construct a three-dimensional network by evolving along the temporal axis.
We refer to this three-dimensional network as the cubic recurrent neural network (CubicRNN). 
Experiment shows that CubicRNN produces highly accurate video predictions on the Moving-MNIST dataset \cite{DBLP:conf/icml/SrivastavaMS15}, Robotic Pushing dataset \cite{DBLP:conf/nips/FinnGL16} and KTH Action dataset \cite{DBLP:conf/icpr/SchuldtLC04}.

\section{Related work}
\subsection{Video Prediction}
A number of prior works have addressed video prediction with different settings. They can essentially be classified as follows.

\textbf{Generation vs. Transformation.} 
As mentioned above, video prediction can be classified as patch level \cite{DBLP:journals/corr/RanzatoSBMCC14}, feature level \cite{DBLP:journals/corr/RanzatoSBMCC14,DBLP:conf/icml/SrivastavaMS15,DBLP:conf/cvpr/VondrickPT16} and image level.
For image-level prediction, the generation group of methods generates each pixel in frames~\cite{DBLP:conf/nips/ShiCWYWW15,DBLP:conf/icml/ReedOKCWCBF17}.
Some of these methods have difficulty in handling real-world video prediction due to the curse of dimensionality.
The second group first predicts a transformation and then applies the transformation to the previous frame to generate a new frame~\cite{DBLP:conf/nips/JiaBTG16,DBLP:conf/nips/JaderbergSZK15,DBLP:conf/nips/FinnGL16,DBLP:conf/cvpr/Vondrick017}.
These types of methods can reduce the difficulty of predicting real-world future frames.
In addition, some methods \cite{DBLP:conf/iclr/VillegasYHLL17,DBLP:conf/nips/DentonB17,tulyakov2017mocogan} first decompose a video into a stationary content part and a temporally varying motion component by multiple loss functions, and then combine the predicted motion and stationary content to construct future frames.

\textbf{Short-term prediction vs. Long-term prediction.} 
Video prediction can be classified as short-term ($<10$ frames) prediction \cite{DBLP:conf/nips/Xue0BF16,DBLP:conf/icml/KalchbrennerOSD17} and long-term ($\geq 10$ frames) prediction \cite{DBLP:conf/nips/OhGLLS15,DBLP:conf/nips/ShiCWYWW15,DBLP:conf/nips/VondrickPT16,DBLP:conf/nips/FinnGL16,DBLP:conf/nips/WangLWGY17} according to the number of predicted frames.
Most methods can usually undertake long-term predictions on virtual-word datasets (e.g., game-play video datasets \cite{DBLP:conf/nips/OhGLLS15}) or synthetic datasets (e.g., the Moving-MNIST dataset \cite{DBLP:conf/icml/SrivastavaMS15}).
Since real-world sequences are less deterministic, some of these methods are limited to making long-term predictions.
For example, dynamic filter network \cite{DBLP:conf/nips/JiaBTG16} is able to predict ten frames on the Moving-MNIST dataset but three frames on the highway driving dataset.
A special case of short-term prediction is the so-called next-frame prediction \cite{DBLP:conf/nips/Xue0BF16,DBLP:conf/iclr/LotterKC17}, which only predicts one frame in the future.

\textbf{Single dependency vs. Multiple dependencies.}
Most methods need to observe multiple context frames before making video predictions.
Other methods, by contrast, aim to predict the future based only on the understanding of a single image \cite{DBLP:conf/cvpr/MottaghiBRF16,DBLP:conf/iclr/MathieuCL15,DBLP:conf/eccv/WalkerDGH16,DBLP:conf/nips/Xue0BF16,DBLP:conf/eccv/ZhouB16,DBLP:conf/cvpr/abs-1709-07592}.

\textbf{Unconditional prediction vs. Conditional prediction.}
The computer vision community usually focuses on unconditional prediction, in which the future only depends on the video itself.
The action-conditional prediction has been widely explored in the robotic learning community, e.g., video game videos \cite{DBLP:conf/nips/OhGLLS15,DBLP:conf/iclr/ChiappaRWM17} and robotic manipulations \cite{DBLP:conf/nips/FinnGL16,DBLP:conf/icml/KalchbrennerOSD17,DBLP:conf/corl/EbertFLL17,DBLP:conf/iclr/Babaeizadeh17,DBLP:conf/icml/ReedOKCWCBF17}.

\textbf{Deterministic model vs. Probabilistic model.} 
One common assumption for video prediction is that the future is deterministic.
Most video prediction methods therefore belong to the deterministic model.
However, the real-world can be full of stochastic dynamics.
The probabilistic model predicts multiple possible frames at one time. \cite{DBLP:conf/iclr/Babaeizadeh17,DBLP:journals/corr/FragkiadakiHAVR17}. 
The probabilistic method is also adopted in single dependency prediction because the precise motion corresponding to a single image is often stochastic and ambiguous \cite{DBLP:conf/nips/Xue0BF16}.


\subsection{Long Short-Term Memory (LSTM)}
LSTM-based methods have been widely used in video prediction \cite{DBLP:conf/nips/FinnGL16,DBLP:conf/nips/WangLWGY17,DBLP:conf/corl/EbertFLL17,DBLP:conf/iclr/Babaeizadeh17,DBLP:conf/iclr/LotterKC17}.  
The proposed CubicLSTM can therefore be considered as a fundamental module for video prediction, which can be applied to many frameworks.

Among multidimensional LSTMs, our CubicLSTM is similar to GridLSTM \cite{DBLP:conf/iclr/KalchbrennerDG15}. 
The difference is that GridLSTM is built by full connections.
Furthermore, all dimensions in GridLSTM are equal.
However, for CubicLSTM, the spatial branch applies $5\times5$ convolutions while the temporal branch applies $1\times1$ convolutions.
Our CubicLSTM is also similar to PyramidLSTM \cite{DBLP:conf/nips/StollengaBLS15}, which consists of six LSTMs along with three axises, to capture the biomedical volumetric image. 
The information flows dependently in the six LSTMs and the output of PyramidLSTM is simply to add the hidden states of the six LSTMs.
However, CubicLSTM has three branches and information flows across the these branches. 
\section{Cubic LSTM}

In this section, we first review Fully-Connected Long Short-Term Memory (FC-LSTM) \cite{DBLP:journals/neco/HochreiterS97} and Convolutional LSTM (ConvLSTM) \cite{DBLP:conf/nips/ShiCWYWW15}, and then describe the proposed Cubic LSTM (CubicLSTM) unit in detail.
\subsection{FC-LSTM}
LSTM is a special recurrent neural network (RNN) unit for modeling long-term dependencies. 
The key to LSTM is the cell state $\mathcal{C}_t$ which acts as an accumulator of the sequence or the temporal information. 
The information from every new input $\mathcal{X}_t$ will be integrated to $\mathcal{C}_t$ if the input $i_t$ is activated.
At the same time, the past cell state $\mathcal{C}_{t-1}$ may be forgotten if the forget gate $f_t$ turns on.
Whether $\mathcal{C}_t$ will be propagated to the hidden state $\mathcal{H}_t$ is controlled by the output gate $o_t$.
Usually, the cell state $\mathcal{C}_t$ and the hidden state $\mathcal{H}_t$ are jointly referred to as the internal LSTM state, denoted as $(\mathcal{C}_t, \mathcal{H}_t)$.
The updates of fully-connected LSTM (FC-LSTM) for the $t$-th time step can be formulated as follows:

\begin{equation}\label{eq:fclstm-1}
\mathrm{FC-LSTM}	\left\{  
             		\begin{array}{lr}

						i_t = \mathrm{\sigma}(\mathcal{W}_i \cdot [\mathcal{X}_t, \mathcal{H}_{t-1}] + b_i),\\
                        
                       	f_t = \mathrm{\sigma}(\mathcal{W}_f \cdot [\mathcal{X}_t, \mathcal{H}_{t-1}] + b_f),\\
                        
                        o_t = \mathrm{\sigma}(\mathcal{W}_o \cdot [\mathcal{X}_t, \mathcal{H}_{t-1}] + b_o), \\
                        
                        c_t = \mathrm{tanh}(\mathcal{W}_c \cdot [\mathcal{X}_t, \mathcal{H}_{t-1}] + b_c), \\
                        
                        \mathcal{C}_t = f_t \odot \mathcal{C}_t + i_t \odot c_t ,\\
                        
                        \mathcal{H}_t = o_t \odot \mathrm{tanh}(\mathcal{C}_t) ,\\
                        
             		\end{array}  
					\right. 
\end{equation}
where $\cdot$, $\odot$ and $[\cdot]$ denote the matrix product, the element-wise product and the concatenation operation, respectively.

To generate $i_t$, $f_t$, $o_t$ and $c_t$ in the FC-LSTM, a fully-connected layer is first applied on the concatenation of the input $\mathcal{X}_t$ and the last hidden state $\mathcal{H}_{t-1}$ with the form $\mathcal{W} \cdot [\mathcal{X}_t, \mathcal{H}_{t-1}] + b$, where $\mathcal{W} = [\mathcal{W}_i, \mathcal{W}_f, \mathcal{W}_o, \mathcal{W}_c]$ and $b = [b_i, b_f, b_o, b_c]$.
The intermediate result is then split into four parts and passed to the activation functions, i.e., $\mathrm{\sigma}$ and $\mathrm{tanh}$.
Lastly, the new state $(\mathcal{C}_t, \mathcal{H}_t)$ is produced according to the gates and $c_t$ by element-wise product.
In summary, FC-LSTM takes the current input $\mathcal{X}_t$ as input, the previous state $(\mathcal{C}_{t-1}, \mathcal{H}_{t-1})$ and generates the new state $(\mathcal{C}_t, \mathcal{H}_t)$. It is parametrized by $\mathcal{W}$, $b$. 
For simplicity, we reformulate FC-LSTM as follow:
\begin{equation}\label{eq:fclstm-2}
\begin{split}
\mathrm{FC-LSTM:} & \\ (\mathcal{C}_t, \mathcal{H}_t) = & \ \mathrm{LSTM}(\mathcal{X}_t, (\mathcal{C}_{t-1}, \mathcal{H}_{t-1}); \mathcal{W}, b; \cdot).
\end{split}
\end{equation}

The input $\mathcal{X}_t$ and the state $(\mathcal{C}_t, \mathcal{H}_t)$ in FC-LSTM are all one-dimensional vectors, which cannot directly encode spatial information.
Although we can use one-dimensional vectors to represent images by flattening the two-dimensional data (greyscale images) or the three-dimensional data (color images), such operation loses spatial correlations.
\begin{figure*}[t]
\centering
\includegraphics[width=1.0\textwidth]{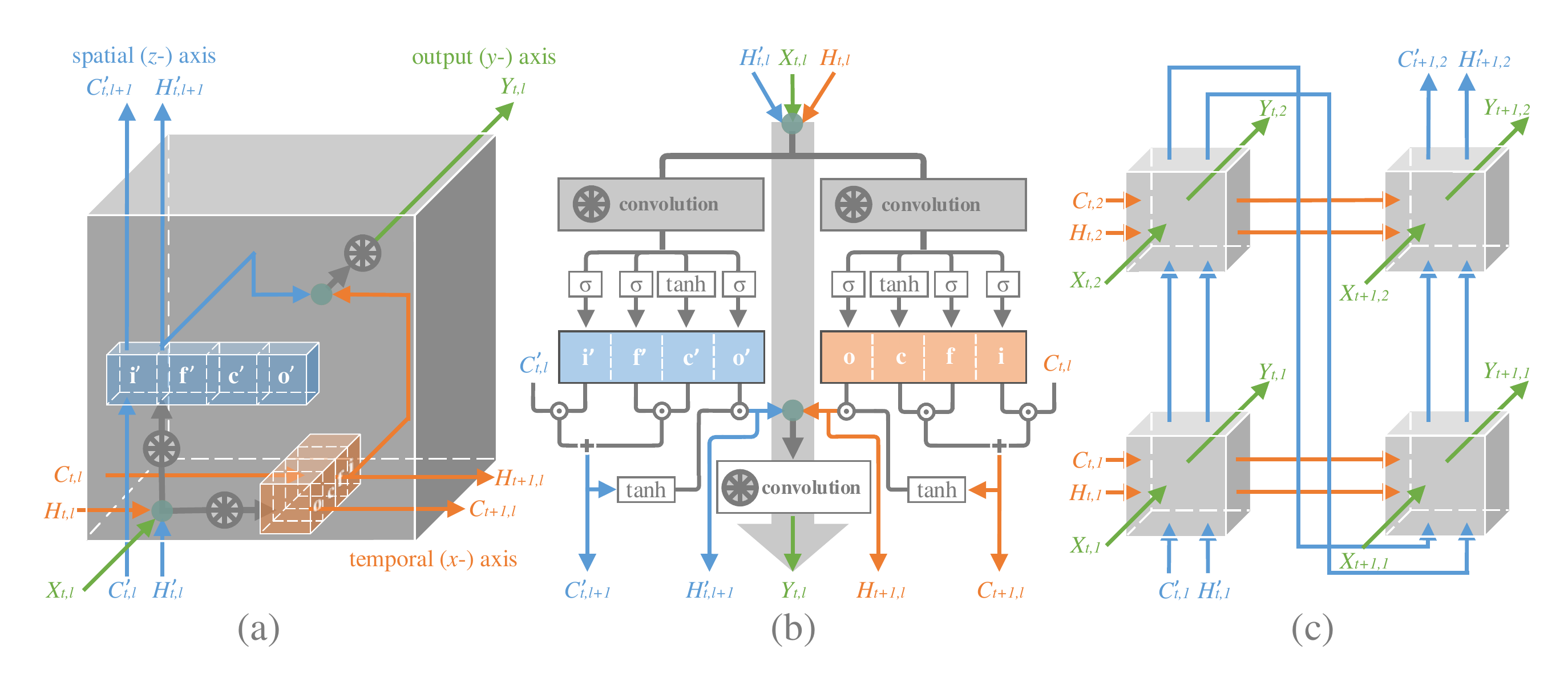}
\caption{(a) 3D structure of the CubicLSTM unit. (b) Topological diagram of the CubicLSTM unit. (c) Two-spatial-layer RNN composed of CubicLSTM units. The unit consists of three branches, a spatial ($z$-) branch for extracting and recognizing moving objects, a temporal ($y$-) branch for capturing and predicting motions, and an output ($x$-) branch for combining the first two branches to generate the predicted frames.
}
\label{cubiclstm}
\end{figure*}

\newcounter{TempEqCnt}  
\setcounter{TempEqCnt}{\value{equation}}
\setcounter{equation}{3}                      
\begin{figure*}[h]
\begin{equation}\label{eq:cubiclstm}
\mathrm{CubicLSTM}	\left\{  
             		\begin{array}{lr}
                    	
                        \mathrm{temporal \ branch:} \ (\mathcal{C}_{t,l}, \mathcal{H}_{t,l}) = \mathrm{LSTM}(\mathcal{X}_{t,l}, \mathcal{H}'_{t,l-1}, (\mathcal{C}_{t-1,l}, \mathcal{H}_{t-1,l}); \mathcal{W}, b; \ast), \\ \\
                        
                        \mathrm{spatial \ branch:} \ (\mathcal{C}'_{t,l}, \mathcal{H}'_{t,l}) = \mathrm{LSTM}(\mathcal{X}_{t,l}, \mathcal{H}_{t-1,l}, (\mathcal{C}'_{t,l-1}, \mathcal{H}'_{t,l-1}); \mathcal{W}', b'; \ast), \\ \\

						\mathrm{output \ branch:} \ \mathcal{Y}_{t,l} = \mathcal{W}'' \ast [\mathcal{H}_{t,l}, \mathcal{H}'_{t,l}] + b''.
						
             		\end{array}  
\right. 
\end{equation}
\end{figure*}
\setcounter{equation}{\value{TempEqCnt}}

\subsection{ConvLSTM}
To exploit spatial correlations for video prediction, ConvLSTM takes three-dimensional tensors as input and replaces the fully-connected layer (matrix product) in FC-LSTM with the convolutional layer.
The updates for ConvLSTM can be written as follow:
\begin{equation}\label{eq:convlstm}
\begin{split}
\mathrm{ConvLSTM:} & \\ (\mathcal{C}_t, \mathcal{H}_t) = & \ \mathrm{LSTM}(\mathcal{X}_t, (\mathcal{C}_{t-1}, \mathcal{H}_{t-1}); \mathcal{W}, b; \ast),
\end{split}
\end{equation}
where $\ast$ denotes the convolution operator and $\mathcal{X}_t$, $(\mathcal{C}_{t-1}, \mathcal{H}_{t-1})$, $(\mathcal{C}_t, \mathcal{H}_t)$ are all three-dimensional tensors with shape (\textit{height}, \textit{width}, \textit{channel}). 
As FC-LSTMs are designed to learn only one type of information, directly adapting FC-LSTM makes it difficult for ConvLSTM to simultaneously process the temporal information and the spatial information in videos.
To predict future spatio-temporal sequences, as previously noted, the convolution in ConvLSTM has to catch motions on one hand and capture moving objects on the other hand.
Similarly, the state of ConvLSTM must be capable of storing motion information and visual information at the same time.

\subsection{CubicLSTM}
To reduce the burden of prediction, the proposed CubicLSTM unit processes the temporal information and the spatial information separately.
Specifically, CubicLSTM consists of three branches: a temporal branch, a spatial branch and a output branch.
The temporal branch aim to obtain and process motions. 
The spatial branch is responsible for capturing and analyzing objects. 
The output branch generates predictions according to the predicted motion information and the moving object information.
As shown in Figure~\ref{cubiclstm}(a), the unit is built along the three axes in a space Cartesian coordinate system.
\begin{itemize}
    \item Along the $x$-axis (temporal axis), the convolution operation obtains the current motion information according to the input $\mathcal{X}_{t,l}$, the previously motion information $\mathcal{H}_{t-1,l}$ and the previously object information $\mathcal{H}'_{t,l-1}$. 
    The current motion information is then used to update the previous temporal cell state $\mathcal{C}_{t-1,l}$ and produce the new motion information $\mathcal{H}_{t,l}$.
    \item Along the $z$-axis (spatial axis), the convolution captures the current spatial layout of objects. 
    This information is then used to rectify the previous spatial cell state $\mathcal{C}'_{t,l-1}$ and generate the new object visual information $\mathcal{H}'_{t,l}$.
    \item Along the $y$-axis (output axis), the output branch combines the current motion information and the current object information to generate an intermediate prediction for the input of the next CubicLSTM unit or construct the final prediction frame. 
\end{itemize}
The topological diagram of ConvLSTM is shown in Figure~\ref{cubiclstm}(b). 
The updates of CubicLSTM are formulated as Eq. (\ref{eq:cubiclstm}), where $l$, $'$ and $''$  denote the spatial network layer, the spatial branch and the output branch respectively.

Essentially, in a single CubicLSTM, the temporal branch and the spatial branch are symmetrical.
However, in experiment, we found that adopting $1 \times 1$ convolution for temporal branch achieves higher accuracy than adopting $5 \times 5$ convolution.
This may indicate that `motion' should focus on temporal neighbors while `object' should focus on spatial neighbors. 
Their functions also depend on their positions and connections in RNNs.
To deal with a sequence, the same unit is repeated along the temporal axis.
Therefore, the parameters are shared along the temporal dimension.
For the spatial axis, we stack multiple different units to form a multi-layer structure to better exploit spatial correlations and capture objects. 
The parameters along the spatial axis are different.

At the end of spatial direction, rather than being discarded, the spatial state is used to initialize the starting spatial state at the next time step.
Formally, the spatial states between two time steps are defined as follows:
\setcounter{equation}{4} 
\begin{equation}\label{eq:spatialstate}
\centering
\mathcal{C}'_{t,1} = \mathcal{C}'_{t-1,L}, \ \ \ \mathcal{H}'_{t,1} = \mathcal{H}'_{t-1,L},
\end{equation}
where $L > 1$ is the number of layers along the spatial axis. 
We demonstrate a 2-spatial-layer RNN in Figure~\ref{cubiclstm}(c), which is the smallest network formed by CubicLSTMs.

\section{Cubic RNN}

\begin{figure}[t]
\centering
\includegraphics[width=0.465\textwidth]{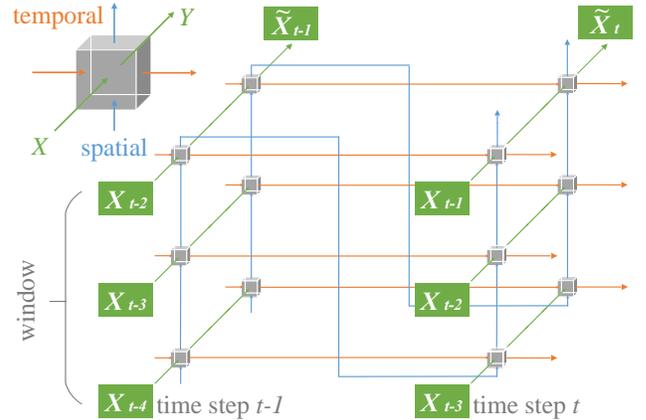}
\caption{A CubicRNN consisting of three spatial layers and two output layers, which can watch three frames at once.
}
\label{cubicrnn}
\end{figure}

In this section, we introduce a new RNN architecture for video prediction, the cubic RNN (CubicRNN).
CubicRNN is created by first stacking multiple CubicLSTM units along the spatial axis and along the output axis, which forms a two-dimensional network, and then evolving along the temporal branch,  which forms a three-dimensional structure. 

In contrast to traditional RNN structures, CubicRNN is capable of watching multiple adjacent frames at one time step along the spatial axis, which forms a sliding window.
The size of the sliding window is equal to the number of spatial layers.
Suppose we have $L$ spatial layers: CubicRNN will view the previous $\mathcal{X}_{t-L+1}, \cdots, \mathcal{X}_{t-1}$ frames to predict the $t$-th frame.
The sliding window enables CubicRNN to better capture the information about both motions and objects.
An example of CubicRNN is illustrated in Figure~\ref{cubicrnn}.

\section{Experiments}

\begin{table*}
\footnotesize
\caption{Results of CubicRNN and state-of-the-art models on the Moving-MNIST dataset. The ``CubicRNN ($c \times y \times z $)'' denotes that the CubicRNN model has $z$ spatial layer(s) and $y$ output layer(s), and the channel size of its state is $c$. We report per-frame mean square error (MSE) and per-frame binary cross-entropy (BCE) of generated frames. Lower MSE or CE means better prediction accuracy.}
\label{tabel:comparison}
\centering
\begin{tabular}{l||cc|c}
\hline

\multirow{2}{*}{Models}     		     								& \multicolumn{2}{c|}{MSE}          	& {BCE}      		\\ \cline{2-4} 

									     								& MNIST-2 				& MNIST-3		& MNIST-2 			\\ \hline 
                                            
FC-LSTM \cite{DBLP:conf/icml/SrivastavaMS15}							& 118.3					& 162.4			& 483.2				\\

CridLSTM \cite{DBLP:conf/iclr/KalchbrennerDG15}				            & 111.6					& 157.8			& 419.5				\\

ConvLSTM ($128 \times 4$) \cite{DBLP:conf/nips/ShiCWYWW15}				& 103.3					& 142.1			& 367.0				\\

PyramidLSTM \cite{DBLP:conf/nips/StollengaBLS15}                        & 100.5					& 142.8			& 355.3				\\

CDNA \cite{DBLP:conf/nips/FinnGL16}										& 97.4					& 138.2			& 346.6				\\

DFN \cite{DBLP:conf/nips/JiaBTG16}										& 89.0					& 130.5			& 285.2				\\

VPN baseline \cite{DBLP:conf/icml/KalchbrennerOSD17} 					& 70.0					& 125.2			& 110.1				\\

PredRNN with spatialtemporal memory \cite{DBLP:conf/nips/WangLWGY17} 	& 74.0 					& 118.2			& 118.5				\\ 

PredRNN + ST-LSTM ($128 \times 4$) \cite{DBLP:conf/nips/WangLWGY17} 	& 56.8 					& 93.4			& 97.0				\\ \hline \hline

CubicRNN ($32 \times 3 \times 1 $)										& 111.5					& 158.4			& 386.3				\\

CubicRNN ($32 \times 1 \times 3 $)										& 73.4					& 127.1			& 210.3				\\ \hline

CubicRNN ($32 \times 3 \times 2 $)										& 59.7					& 110.2			& 121.9				\\

CubicRNN ($32 \times 3 \times 3 $)										& \textbf{47.3}			& \textbf{88.2}	& \textbf{91.7}				\\ \hline 
\end{tabular}
\end{table*}

We evaluated the proposed CubicLSTM unit on three video prediction datasets, Moving-MNIST dataset \cite{DBLP:conf/icml/SrivastavaMS15}, Robotic Pushing dataset \cite{DBLP:conf/nips/FinnGL16} and KTH Action dataset \cite{DBLP:conf/icpr/SchuldtLC04}, including synthetic and real-world video sequences.
All models were trained using the ADAM optimizer \cite{DBLP:conf/iclr/KingmaB14} and implemented in TensorFlow. 
We trained the models using eight GPUs in parallel and set the batch size to four for each GPU.

\subsection{Moving MNIST dataset}
The Moving MNIST dataset consists of 20 consecutive frames, 10 for the input and 10 for the prediction.
Each frame contains two potentially overlapping handwritten digits moving and bouncing inside a $64 \times 64$ image.
The digits are chosen randomly from the MNIST dataset and placed initially at random locations.
Each digit is assigned a velocity whose direction is chosen uniformly on a unit circle and whose magnitude is also chosen uniformly at random over a fixed range.
The size of the training set can be considerable large.
We use the code provided by \cite{DBLP:conf/nips/ShiCWYWW15} to generate training samples on-the-fly.
For the test, we followed \cite{DBLP:conf/nips/WangLWGY17} which evaluates methods on two settings, i.e., two moving digits (MNIST-2) and three moving digits (MNIST-3).
We also followed \cite{DBLP:conf/nips/WangLWGY17} to evaluate predictions, in which both the mean square error (MSE) and binary cross-entropy (BCE) were used. 
Only simple prepossessing was done to convert pixel values into the range $[0,1]$.

Each state in the implementation of CubicLSTM has 32 channels.
The size of the spatial-convolutional kernel was set to $5 \times 5$.
Both the temporal-convolutional kernel and output-convolutional kernel were set to $1 \times 1$.
We used the MSE loss and BCE loss to train the models corresponding to the different evaluation metrics.
We also adopted an encoder-decoder framework as \cite{DBLP:conf/icml/SrivastavaMS15,DBLP:conf/nips/ShiCWYWW15}, in which the initial states of the decoder network are copied from the last states of the encoder network.
The inputs and outputs were fed and generated as CubicRNN (Figure \ref{cubicrnn}).
All models were trained for $300K$ iterations with a learning rate of $10^{-3}$ for the first $150K$ iterations and a learning rate of $10^{-4}$ for the latter $150K$ iterations.

\begin{figure}[t]
\centering
\includegraphics[width=0.4725\textwidth]{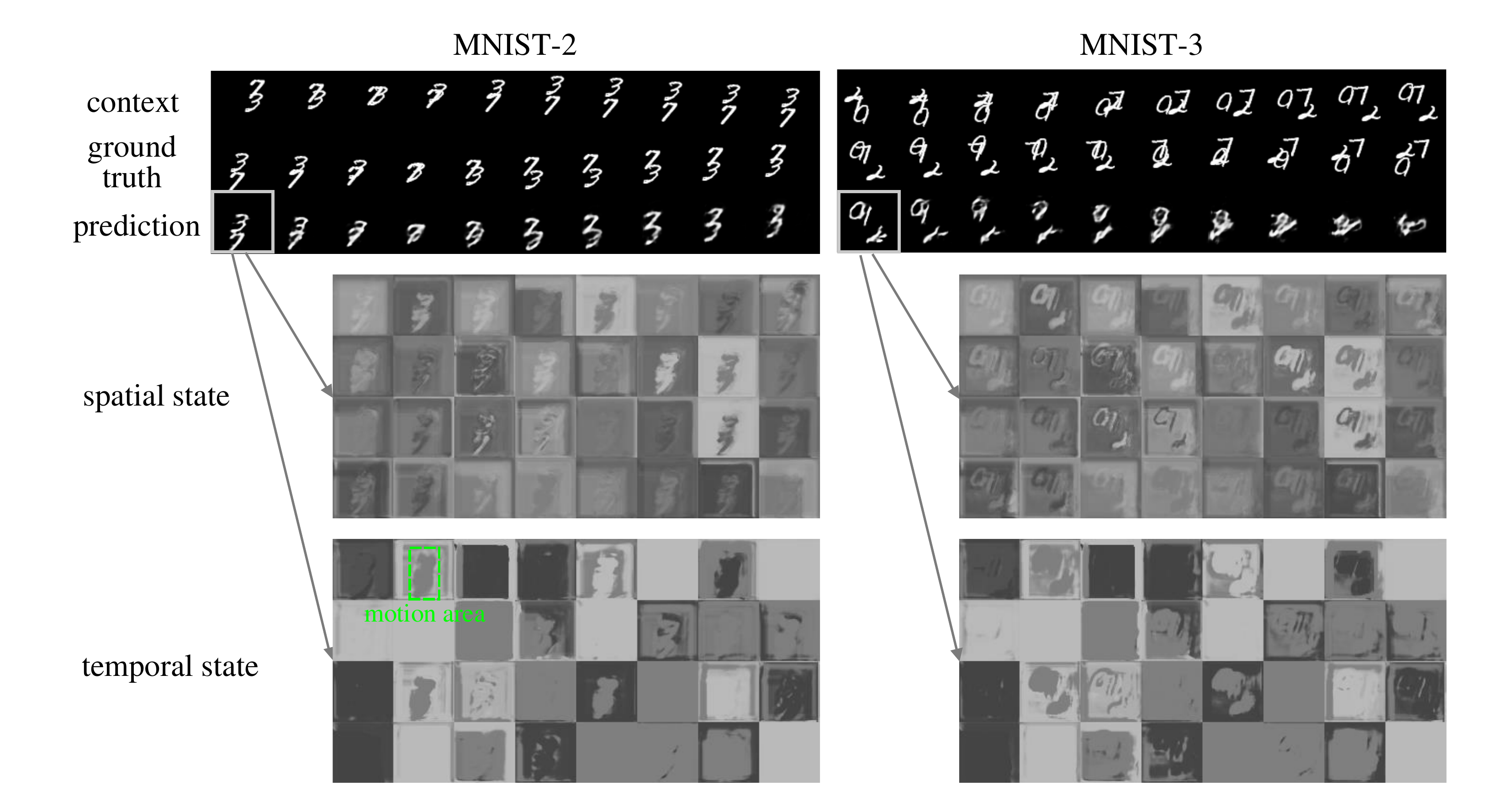}
\caption{Prediction examples on the Moving MNIST dataset (top) and visualizations of the spatial hidden state (middle) and temporal hidden state (bottom).
The spatial state provides the object information such as contours and appearances, while the temporal state provides the motion information of potential motion areas. 
The two states will be exploited to generate the future frame by the output branch.}
\label{mmnist}
\end{figure}

\textbf{Improvement by the spatial branch.}
Compared to ConvLSTM \cite{DBLP:conf/nips/ShiCWYWW15}, CubicLSTM has an additional spatial branch.
To prove the improvement achieved by this branch, we compare two models, both of which have 3 CubicLSTM units.
The first model stacks units along the output axis forms, forming a structure of 1 spatial layer and 3 output layers.
Since each state has 32 channels, we denote this structure as $(32 \times 3 \times 1)$.
The second model stacks the three units along the spatial axis, forming a structure of 3 spatial layers and 1 output layer structure, denoted as $(32 \times 1 \times 3)$.
The results are listed in Table~\ref{tabel:comparison}.

\begin{figure*}[t]
\centering
\includegraphics[width=0.985\textwidth]{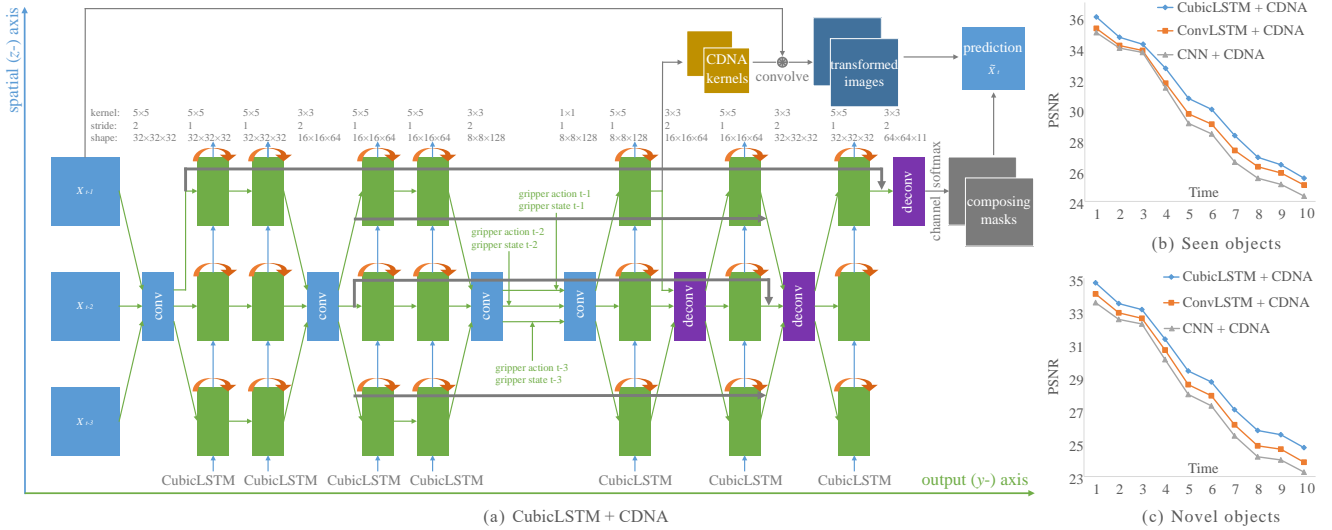}
\caption{(a): Architecture of our model for the Robotic Pushing dataset. 
The model is largely borrowed from the convolutional dynamic neural advection (CDNA) model proposed in \cite{DBLP:conf/nips/FinnGL16} and replaces ConvLSTMs in the CNDA model with CubicLSTMs.
Our model has three spatial layers, among which the convolutions and the deconvolutions are shared.
(b)-(c): Frame-wise PSNR comparisons of ``CubicLSTM + CDNA'', ``ConvLSTM + CDNA'' \cite{DBLP:conf/nips/FinnGL16} and ``CNN + CDNA'' on the Robotic Pushing dataset. Higher PSNR means better prediction accuracy.}
\label{pushing}
\end{figure*}

On one hand, although both have 3 CubicLSTM units, CubicRNN $(32 \times 1 \times 3)$ significantly outperforms CubicRNN $(32 \times 3 \times 1)$. 
As the spatial branch and the temporal branch in CubicRNN $(32 \times 3 \times 1)$ are identical, the model does not exploit spatial correlations well and only achieves similar accuracy to ConvLSTM $(128 \times 4)$.
On the other hand, even though CubicRNN $(32 \times 1 \times 3)$ uses fewer parameters than ConvLSTM $(128 \times 4)$, it obtains better predictions.
This experiment validates our belief that the temporal information and the spatial information should be processed separately.

\textbf{Comparison with other models.} 
We report all the results from existing models on the Moving MNIST dataset (Table~\ref{tabel:comparison}). 
We report two CubicRNN settings. 
The first has three output layers and two spatial layers $(32 \times 3 \times 2)$, and the second one has three output layers and three spatial layers $(32 \times 3 \times 3)$. 
CubicRNN $(32 \times 3 \times 3)$ produces the best prediction.
The prediction of CubicRNN is improved by increasing the number of spatial layers.


\begin{figure*}[t]
\centering
\includegraphics[width=1.0\textwidth]{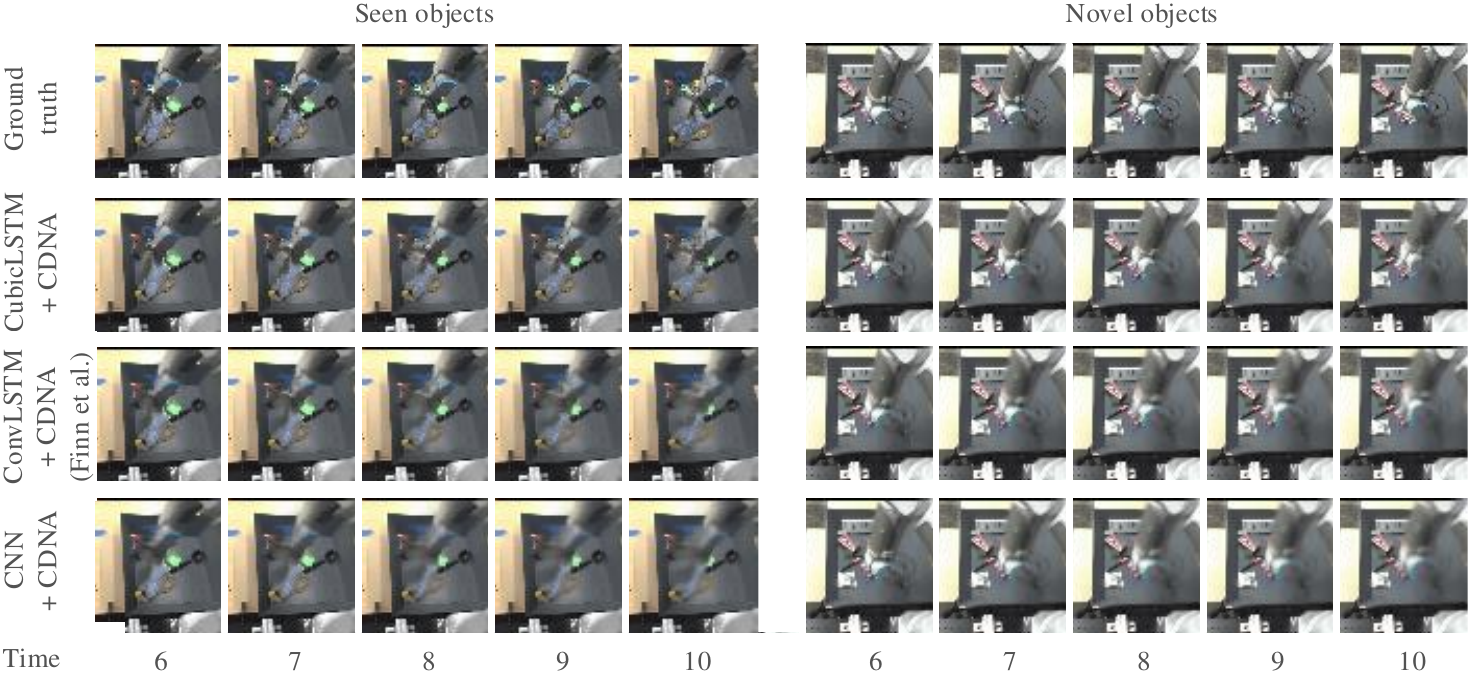}
\caption{Qualitative comparisons of ``CubicLSTM + CDNA'', ``ConvLSTM + CDNA'' \cite{DBLP:conf/nips/FinnGL16} and ``CNN + CDNA'' on the Robotic Pushing dataset. Our ``CubicLSTM + CDNA'' method can generate clearer frames than others, especially for the videos with novel objects.}
\label{pushvis}
\end{figure*}

We illustrate two prediction examples produced by CubicRNN $(32 \times 3 \times 3)$ in the first row of Figure~\ref{mmnist}.
The model is capable of generating accurate predictions for the two-digit case (MNIST-2).
In the second and third rows of Figure~\ref{mmnist}, we visualize the spatial hidden states and the temporal states of the last CubicLSTM unit in the $(32 \times 3 \times 3)$ structure when it predicts the first frames of the two prediction examples. 
These states have 32 channels and we visualize each channel by the function $\mathrm{\sigma}(\cdot) \times 255$.
As can be seen from the visualization, the spatial hidden state reflects the visual and the spatial information of the digits in the frame.  
The temporal hidden state suggests that it is likely to contain some ``motion areas''.
Compared with the relatively precise information of objects provided by the spatial hidden state, the ``motion areas'' provided by the temporal hidden state are somewhat rough.
The output branch will apply these ``motion areas'' on the digits to generate the prediction.

\subsection{Robotic Pushing dataset}

Robotic Pushing~\cite{DBLP:conf/nips/FinnGL16} is an action-conditional video prediction dataset which recodes 10 robotic arms pushing hundreds of objects in a basket.
The dataset consists of 50,000 iteration sequences with 1.5 million video frames and two test sets.
Objects in the first test set use two subsets of objects in the training set, which are so-called ``seen objects''.
The second test set involves two subsets of ``novel objects'', which are not used during training.
Each test set has 1,500 recoded sequences.
In addition to RGB images, the dataset also provides the corresponding gripper poses by recoding its states and commanded actions, both of which are 5-dimensional vectors.
We follow \cite{DBLP:conf/nips/FinnGL16} to center-crop and downsample images to $64 \times 64$, and use the Peak Signal to Noise Ratio (PSNR) \cite{DBLP:conf/iclr/MathieuCL15} to evaluate the prediction accuracy.

Our model is illustrated in Figure~\ref{pushing}(a).
The model is largely borrowed from the convolutional dynamic neural advection (CDNA) model \cite{DBLP:conf/nips/FinnGL16} which is an encoder-decoder architecture that expands along the output direction and consists of several convolutional encoders, ConvLSTMs, deconvolutional decoders and CDNA kernels.
We denote the CDNA model in \cite{DBLP:conf/nips/FinnGL16} as ``ConvLSTM + CDNA''.
Our model replaces ConvLSTMs in the CNDA model with CubicLSTMs and expands the model along the spatial axis to form a three-spatial-layer architecture.
The convolutional encoders and the deconvolutional decoders are shared among these three spatial layers.
We refer to our model as ``CubicLSTM + CDNA''.
We also design a baseline model which replaces ConvLSTMs in the CNDA model with CNNs, denoted as ``CNN + CDNA''.
Since CNNs do not have internal state that flows along the temporal dimension, it cannot exploit the temporal information.
In this experiment, all models were given three context frames before predicting 10 future frames and were trained for $100K$ iterations with the mean square error loss and the learning rate of $10^{-3}$.
The results are shown in Figure~\ref{pushing}(b) and Figure~\ref{pushing}(c). 

The ``ConvLSTM + CDNA'' model only obtains similar accuracy to the ``CNN + CDNA'' model, which indicates that convolutions of ConvLSTMs in the CDNA model may mainly focus on the spatial information while neglecting the temporal information. 
The ``CubicLSTM + CDNA'' model achieves the best prediction accuracy. 
A qualitative comparison of predicted video sequences is given in Figure~\ref{pushvis}.
The ``CubicLSTM + CDNA'' model generates sharper frames than other models.

\subsection{KTH Action dataset}

The KTH Action dataset~\cite{DBLP:conf/icpr/SchuldtLC04} is a real-world videos of people performing one of six actions (walking, jogging, running, boxing, handwaving, hand-clapping) against fairly uniform backgrounds.
We compared our method with the Motion-Content Network (MCnet) \cite{DBLP:conf/iclr/VillegasYHLL17} and Disentangled Representation Net (DrNet) \cite{DBLP:conf/nips/DentonB17} on the KTH Action dataset.

\begin{figure}[h]
\centering
\includegraphics[width=0.465\textwidth]{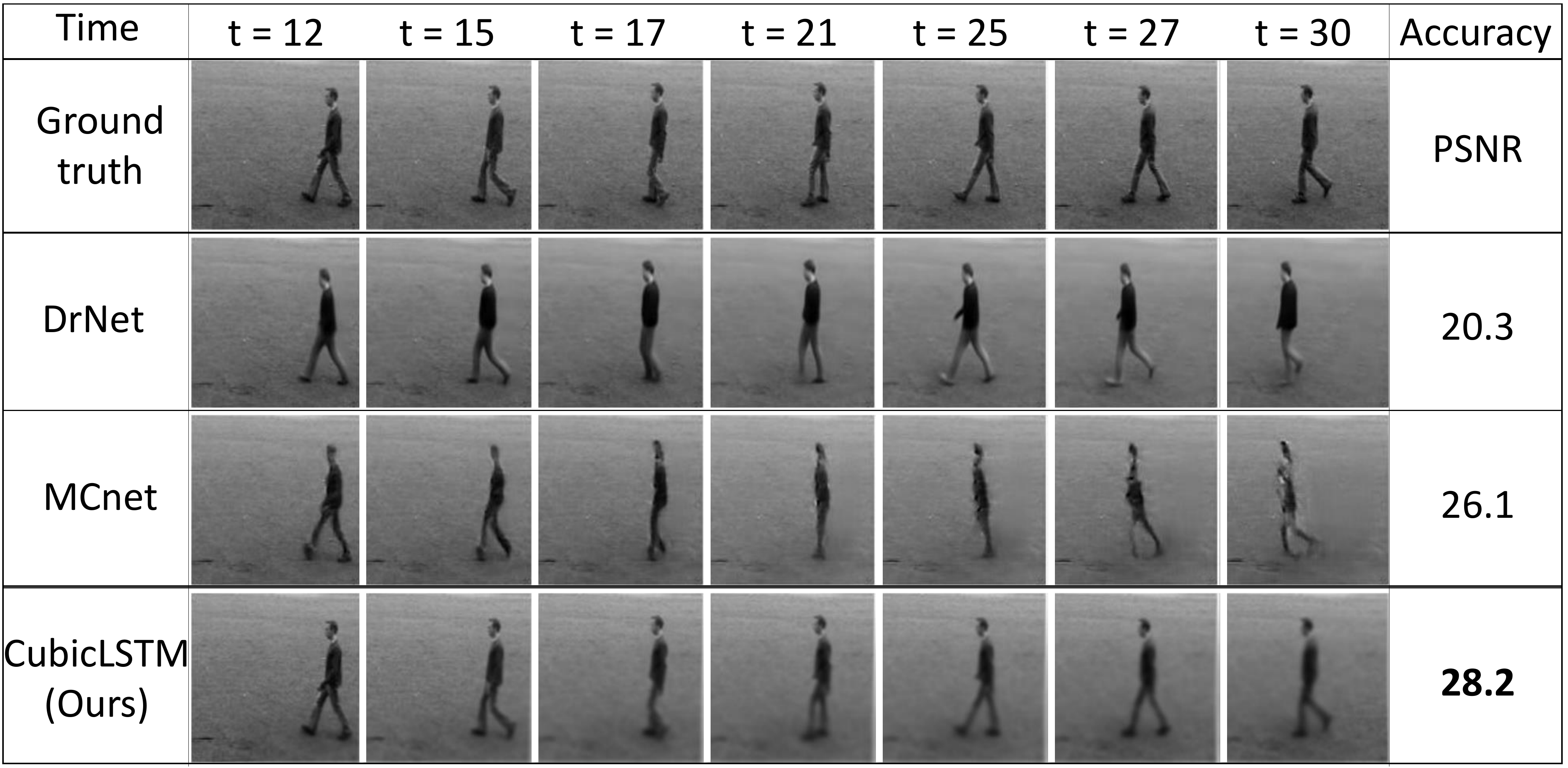}
\caption{Qualitative and quantitative comparison of generated sequences between DrNet, MCnet and CubicLSTM.}
\label{kth}
\end{figure}

The qualitative and quantitative comparisons between DrNet, MCnet and CubicLSTM are shown in Figure~\ref{kth}.
Although DrNet can generate relatively clear frames, the appearance and position of the person is slightly changed. Therefore, the PSNR accuracy is not quite high. 
For MCnet, the generated frames are a little distorted.
Compared to DrNet and MCnet, our CubicLSTM model can predict more accurate frames and therefore achieves the highest accuracy. 

\section{Conclusions}

In this work, we develop a CubicLSTM unit for video prediction.
The unit processes spatio-temporal information separately by a spatial branch and a temporal branch.
This separation can reduce the video prediction burden for networks. 
The CubicRNN is created by stacking multiple CubicLSTMs and generates better predictions than prior models, which validates out belief that the spatial information and temporal information should be processed separately.

\clearpage
{
\bibliographystyle{aaai}
\bibliography{egbib}
}

\end{document}